Feedback Linearization Based Tracking Control of A Tilt-rotor with Cat-trot Gait Plan

Zhe Shen, Yudong Ma, and Takeshi Tsuchiya

Abstract

With the introduction of the laterally bounded forces, the tilt-rotor gains more flexibility in the controller design. Typical feedback linearization methods utilize all the inputs in controlling this vehicle; the magnitudes as well as the directions of the thrusts are maneuvered simultaneously based on a unified control rule. Although several promising results indicate that these controllers may track the desired complicated trajectories, the tilting angles are required to change relatively fast or in large scale during the flight, which turns to be a challenge in application. The recent gait plan for a tilt-rotor may solve this problem; the tilting angles are fixed or vary in a predetermined pattern without being maneuvered by the control algorithm. Carefully avoiding the singular decoupling matrix, several attitudes can be tracked without changing the tilting angles frequently. While the position was not directly regulated in that research, which left the position-tracking still an open question. In this research, we elucidate the coupling relationship between the position and the attitude. Based on this, we design the position-tracking controller, adopting feedback linearization. A cat-trot gait is further designed for a tilt-rotor to track the reference; three types of references are designed for our tracking experiments: setpoint, uniform rectilinear motion, and uniform circular motion. The significant improvement with less steady state error is witnessed after equipping with our modified attitude-position decoupler. It is also found that the frequency of the cat-trot gait highly influenced the steady state error.

1. Introduction

Comparing with the conventional quadrotor [1–4], the tilt-rotors [5–9] provides the lateral forces, which are not applicable to the collinear/coplanar platforms (e.g., conventional quadrotors). The additional mechanical structures (usually tilting motors) mounted on the arms of the tilt-rotor provide the possibility of changing the direction of each thrust or 'tilting'. As a consequence, the number of inputs increases to eight (four magnitudes of the thrust and four directions of the thrusts).

One of the systematic methods to solve tracking problem for a conventional quadrotor is feedback linearization (dynamic inversion), which transfers the nonlinear system to a linear one applicable to implement the linear controllers. Hereon, each output of interest is manipulated individually. Since the number of the inputs in conventional quadrotors (four) is less than the number of degrees of freedom (six), a controller can independently stabilize four outputs at most. Typical choices of these four outputs can be attitude-altitude [10–14] and position-yaw [6,15–18].

On the contrary, since the number of inputs in Ryll's tilt-rotors (eight) is larger than the number of degrees of freedom, the vehicle becomes an over-actuated system. This property intrigues the research on fully tracking the entire degrees of freedom; the tilt-rotor [7] not only tracks the desired time-specified position but also the desired attitude relying on the eight inputs [19,20]. Although the promising results yield acceptable tracking result, the tilting angles vary greatly or over-rapidly, which sharpens the feasibility of implementing the relevant controllers.

To solve this problem, our previous research [21] plans the gait of the tilt-rotor before applying feedback linearization. The tilting angles are assigned beforehand, rather than manipulated by a unified control rule. Hereby, the magnitudes of the thrusts are the only inputs.

However, several challenges may hinder the application of this method. One of them is the singular decoupling matrix [22] of feedback linearization. Considerable attention has been paid to this issue for the conventional quadrotor [10,23]. It has been proved that the decoupling matrix is always invertible for a tilt-rotor with an over-actuated control scenario [7]. However, this matrix may be singular if we take the magnitudes of the thrusts as the only inputs after the gait plan for a tilt-rotor. Further discussions on avoiding the singular decoupling matrix can refer to [21].

On the other hand, though [21] witnessed promising result in tracking attitude and altitude, the position $(X,Y)$ is not successfully stabilized. One may notice that the relationship between the acceleration $(\ddot{X},\ddot{Y})$ and the attitude has been elucidated by an attitude-position decoupler in a conventional quadrotor [12,14,20,24–26]. While this decoupler does not hold for a tilt-rotor. A modified attitude-position decoupler applicable to a tilt-rotor is deduced in this article.

Similar to the gait plan problem in four-legged robots [27–29], [21] proposed several gaits for a tilt-rotor, averting the singular decoupling matrix. However, the gait in [21] did not consider the gait patterns with varying tilting angles; all the gaits analyzed were the combinations of the fixed tilting angles. Another contribution of this research is to adopt the time-specified varying gait, which is inspired by cat trot [30–32].

Three types of references are designed to track for a tilt-rotor in simulation by Simulink. Notable improvements in tracking result are witnessed after the application with our modified attitude-position decoupler.

The rest of the article is structured as follows: Section 2 briefs the dynamics of the tilt-rotor. Section 3 explores the relationship coupled in attitude and position in a tilt-rotor before proposing the modified attitude-position decoupler, which is further used to design a controller. Section 4 introduces a gait for a tilt-rotor which is inspired by a cat-trot gait. The discussions upon the stability of the relevant controller is addressed in Section 5. In Section 6 introduces the settings of the simulation tests, the results of which are displayed in Section 7. Finally, we make conclusions and discussions in Section 8.

2. Dynamics of The Tilt-rotor

Figure 1 [21] sketches Ryll's tilt-rotor. This tilt-rotor [7,21] can adjust the direction of each thrust during the flight; tilting each arm changes the direction of the thrust, which is restricted in the relevant yellow plane in Figure 1. $\alpha_i$ ($i$=1,2,3,4) represents the tilting angles.

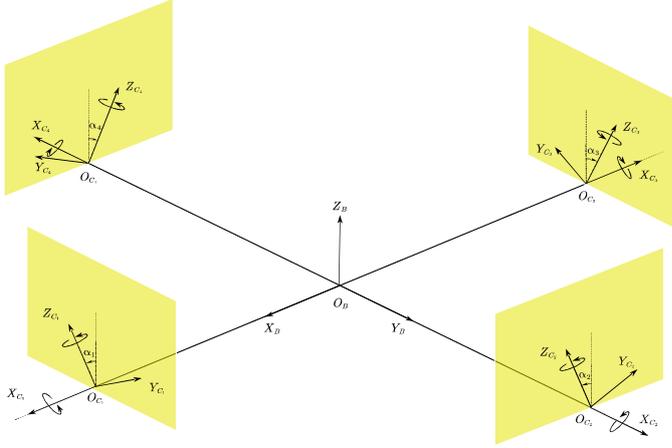

Figure 1. The sketch of the tilt-rotor.

The position $P = [X \quad Y \quad Z]^T$ is ruled by

$$\ddot{P} = \begin{bmatrix} 0 \\ 0 \\ -g \end{bmatrix} + \frac{1}{m} \cdot {}^W R \cdot F(\alpha) \cdot \begin{bmatrix} \varpi_1 \cdot |\varpi_1| \\ \varpi_2 \cdot |\varpi_2| \\ \varpi_3 \cdot |\varpi_3| \\ \varpi_4 \cdot |\varpi_4| \end{bmatrix} \triangleq \begin{bmatrix} 0 \\ 0 \\ -g \end{bmatrix} + \frac{1}{m} \cdot {}^W R \cdot F(\alpha) \cdot w \qquad (1)$$

where $m$ is the total mass, $g$ is the gravitational acceleration, $\varpi_i$, $(i = 1,2,3,4)$ is the angular velocity of the propeller ($\varpi_{1,3} < 0, \varpi_{2,4} > 0$) with respect to the propeller-fixed frame, ${}^W R$ is the rotational matrix [4] from the inertial frame to the body-fixed frame,

$$ {}^W R = \begin{bmatrix} c\theta \cdot c\psi & s\phi \cdot s\theta \cdot c\psi - c\phi \cdot s\psi & c\phi \cdot s\theta \cdot c\psi + s\phi \cdot s\psi \\ c\theta \cdot s\psi & s\phi \cdot s\theta \cdot s\psi + c\phi \cdot c\psi & c\phi \cdot s\theta \cdot s\psi - s\phi \cdot c\psi \\ -s\theta & s\phi \cdot c\theta & c\phi \cdot c\theta \end{bmatrix} \qquad (2)$$

where $s\Lambda = \sin(\Lambda)$, $c\Lambda = \cos(\Lambda)$, $\phi$, $\theta$, and $\psi$ are roll angle, pitch angle, and yaw angle, respectively, tilting angles $\alpha = [\alpha_1 \quad \alpha_2 \quad \alpha_3 \quad \alpha_4]$, the positive directions of which are defined in Figure 1, $F(\alpha)$ is defined as

$$F(\alpha) = \begin{bmatrix} 0 & K_f \cdot s2 & 0 & -K_f \cdot s4 \\ K_f \cdot s1 & 0 & -K_f \cdot s3 & 0 \\ -K_f \cdot c1 & K_f \cdot c2 & -K_f \cdot c3 & K_f \cdot c4 \end{bmatrix} \qquad (3)$$

where $si = \sin(\alpha_i)$, $ci = \cos(\alpha_i)$, $(i = 1,2,3,4)$. $K_f$ ($8.048 \times 10^{-6} N \cdot s^2/rad^2$) is the coefficient of the thrust.

The angular velocity of the body with respect to its own frame, $\omega_B = [p \quad q \quad r]^T$, is governed by

$$\dot{\omega}_B = I_B^{-1} \cdot \tau(\alpha) \cdot w \qquad (4)$$

where $I_B$ is the matrix of moments of inertia, $K_m$ ($2.423 \times 10^{-7} N \cdot m \cdot s^2/rad^2$) is the coefficient of the drag moment, $L$ is the length of the arm, $\tau(\alpha)$ is defined by

$$\tau(\alpha) = \begin{bmatrix} 0 & L \cdot K_f \cdot c2 - K_m \cdot s2 & 0 & -L \cdot K_f \cdot c4 + K_m \cdot s4 \\ L \cdot K_f \cdot c1 + K_m \cdot s1 & 0 & -L \cdot K_f \cdot c3 - K_m \cdot s3 & 0 \\ L \cdot K_f \cdot s1 - K_m \cdot c1 & -L \cdot K_f \cdot s2 - K_m \cdot c2 & L \cdot K_f \cdot s3 - K_m \cdot c3 & -L \cdot K_f \cdot s4 - K_m \cdot c4 \end{bmatrix}.$$
(5)

The relationship between the angular velocity of the body, $\omega_B$, and the attitude rotation matrix ($^W R$) is given [33–35] by

$$^W\dot{R} = {^W R} \cdot \widehat{\omega}_B \quad (6)$$

where " $\widehat{\phantom{x}}$ " is the hat operation to produce the skew matrix.

The parameters of this tilt-rotor are specified as follows: $m = 0.429 kg$, $L = 0.1785 m$, $g = 9.8 N/kg$, $I_B = \text{diag}([2.24 \times 10^{-3}, 2.99 \times 10^{-3}, 4.80 \times 10^{-3}]) kg \cdot m^2$.

## 3. Attitude-position Decoupler and Feedback Linearization

The controller for the tilt-rotor comprises three parts: modified attitude-position decoupler, feedback linearization, and high-order PD controller.

### 3.1. Scenario of The Controller

As mentioned, this controller can only control four degrees of freedom independently at most. Noticing that picking position and yaw may introduce the singular decoupling matrix [10,23], the independently controlled variables decided in this research are attitude and altitude [11,12,20,25].

The rest of the degrees of freedom (position ($X,Y$)) are tracked by adjusting the attitude, relying on the coupled relationship between the position and attitude (Section 3.2). This nested structure can be found in Figure 2 (green part).

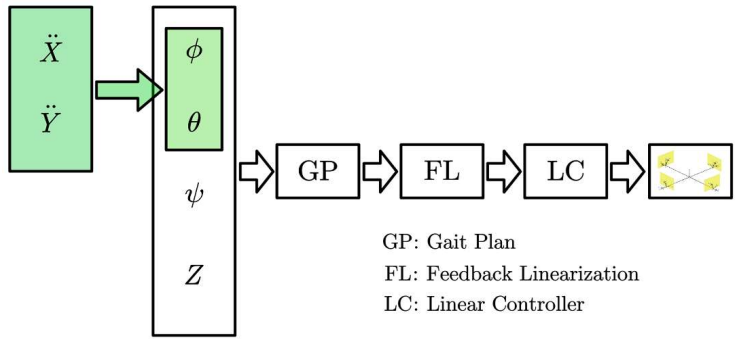

Figure 2. The Control diagram.

After dealing with the coupled position and attitude, an animal-inspired gait is designed for the tilt-rotor, which is detailed in Section 4.

Finally, feedback linearization (Section 3.3) is applied to accommodate a linear controller (Section 3.4).

### 3.2. Modified Attitude-position Decoupler

The relationship (conventional attitude-position decoupler) between attitude and position in a conventional quadrotor [10–14,17,20,25,26,36–38] is

$$\phi = \frac{1}{g} \cdot \left( \ddot{X} \cdot s\psi - \ddot{Y} \cdot c\psi \right), \tag{7}$$

$$\theta = \frac{1}{g} \cdot \left( \ddot{X} \cdot c\psi + \ddot{Y} \cdot s\psi \right). \tag{8}$$

This relationship is deduced by linearizing the dynamics of a conventional quadrotor near hovering. However, it attitude-position decoupler does not hold for the tilt-rotor if the tilting angles are not all zero. Proposition 1 gives the modified attitude-position decoupler for a tilt-rotor.

**Proposition 1 (Modified Attitude-position Decoupler)**

The attitude and the position of a tilt-rotor can be approximately decoupled as

$$\phi = \frac{1}{g} \cdot \left( \ddot{X} \cdot s\psi - \ddot{Y} \cdot c\psi \right) + \frac{F_Y}{mg}, \tag{9}$$

$$\theta = \frac{1}{g} \cdot \left( \ddot{X} \cdot c\psi + \ddot{Y} \cdot s\psi \right) - \frac{F_X}{mg}, \tag{10}$$

where $F_X$ and $F_Y$ are defined by

$$\begin{bmatrix} F_X \\ F_Y \end{bmatrix} = K_f \cdot \begin{bmatrix} 0 & s2 & 0 & -s4 \\ s1 & 0 & -s3 & 0 \end{bmatrix} \cdot \begin{bmatrix} I_B^{-1} \cdot \tau(\alpha) \\ \frac{K_f}{m} \cdot [0 \ 0 \ 1] \cdot F(\alpha) \end{bmatrix}^{-1} \cdot \begin{bmatrix} 0 \\ 0 \\ 0 \\ g \end{bmatrix} \tag{11}$$

**Proof**

Linearize Equation (1) at the equilibrium states,

$$\dot{\omega}_B = I_B^{-1} \cdot \tau(\alpha) \cdot w = 0, \tag{12}$$

$$\ddot{Z} = 0, \tag{13}$$

$$\phi, \theta = 0. \tag{14}$$

Ignoring the high-order infinitesimal terms (e.g., $\phi \cdot \theta, \phi^2, \theta^2$) yields Equation (11).

**Remark 2**

Given $\alpha_1 = \alpha_2 = \alpha_3 = \alpha_4 = 0$, we receive $F_X = F_Y = 0$. Subsequently, Equation (9), (10) degrade to Equation (7), (8) in this special case. In other words, the attitude-position decoupler for the conventional quadrotor is a special case of the modified attitude-position decoupler for the tilt-rotor.

3.3. Feedback Linearization

The degrees of freedoms for the controller to tracked independently are attitude and altitude.

Define

$$\begin{bmatrix} y_1 \\ y_2 \\ y_3 \\ y_4 \end{bmatrix} = \begin{bmatrix} \phi \\ \theta \\ \psi \\ Z \end{bmatrix}. \qquad (15)$$

Since $\varpi_{1,3} < 0$, $\varpi_{2,4} > 0$, we have

$$(\varpi_i \cdot |\varpi_i|)' = 2 \cdot \dot\varpi_i \cdot |\varpi_i|. \qquad (16)$$

Assuming

$$\dot\alpha_i \equiv 0, i = 1,2,3,4, \qquad (17)$$

calculating the third derivative of Equation (18) [21] yields

$$\begin{bmatrix} \dddot y_1 \\ \dddot y_2 \\ \dddot y_3 \\ \dddot y_4 \end{bmatrix} = \begin{bmatrix} I_B^{-1} \cdot \tau(\alpha) \\ [0\ \ 0\ \ 1] \cdot \frac{K_f}{m} \cdot {}^WR \cdot F(\alpha) \cdot 2 \cdot \begin{bmatrix} |\varpi_1| & & & \\ & |\varpi_2| & & \\ & & |\varpi_3| & \\ & & & |\varpi_4| \end{bmatrix} \end{bmatrix}^{4\times 4} \cdot \begin{bmatrix} \dot\varpi_1 \\ \dot\varpi_2 \\ \dot\varpi_3 \\ \dot\varpi_4 \end{bmatrix} + [0\ \ 0\ \ 1] \cdot \frac{K_f}{m} \cdot {}^WR \cdot$$

$$\hat\omega_B \cdot F(\alpha) \cdot w \cdot \begin{bmatrix} 0 \\ 0 \\ 0 \\ 1 \end{bmatrix} \triangleq \bar\Delta \cdot \begin{bmatrix} \dot\varpi_1 \\ \dot\varpi_2 \\ \dot\varpi_3 \\ \dot\varpi_4 \end{bmatrix} + Ma \qquad (18)$$

where $\bar\Delta$ is called decoupling matrix [22]. $[\dot\varpi_1\ \ \dot\varpi_2\ \ \dot\varpi_3\ \ \dot\varpi_4]^T \triangleq U$ is the updated input vector.

Observing Equation (18), one may receive the decoupled relationship in Equation (19), which is compatible with the controller design process, which will be detailed in Section 3.4.

$$\begin{bmatrix} \dot\varpi_1 \\ \dot\varpi_2 \\ \dot\varpi_3 \\ \dot\varpi_4 \end{bmatrix} = \bar\Delta^{-1} \cdot \left( \begin{bmatrix} \dddot y_{1d} \\ \dddot y_{2d} \\ \dddot y_{3d} \\ \dddot y_{4d} \end{bmatrix} - Ma \right). \qquad (19)$$

Obviously, it should be guaranteed that the decoupling matrix ($\bar\Delta$) in Equation (19) is invertible. [21] details the equivalent requirements for receiving an invertible decoupling matrix in Proposition 2.

**Proposition 2.** When the roll angle and pitch angle of the tilt-rotor are close to zero, the decoupling matrix is invertible if and only if Condition (20) holds.

$4.000 \cdot c1 \cdot c2 \cdot c3 \cdot c4 + 5.592 \cdot (+c1 \cdot c2 \cdot c3 \cdot s4 - c1 \cdot c2 \cdot s3 \cdot c4 + c1 \cdot s2 \cdot c3 \cdot c4 - s1 \cdot c2 \cdot c3 \cdot c4) + 0.9716 \cdot (+c1 \cdot c2 \cdot s3 \cdot s4 + c1 \cdot s2 \cdot s3 \cdot c4 + s1 \cdot c2 \cdot s3 \cdot s4 + s1 \cdot s2 \cdot c3 \cdot s4) + 2.000 \cdot (-c1 \cdot s2 \cdot c3 \cdot s4 - s1 \cdot c2 \cdot s3 \cdot c4) + 0.1687 \cdot (-c1 \cdot s2 \cdot s3 \cdot s4 + s1 \cdot c2 \cdot s3 \cdot s4 - s1 \cdot s2 \cdot c3 \cdot s4 + s1 \cdot s2 \cdot s3 \cdot c4) \neq 0 \qquad (20)$

**Proof**

See [21].

In this research, we adopt the cat-trot gait defined by

$$\alpha_1 = -\rho, \qquad (21)$$

$$\alpha_2 = -\rho, \tag{22}$$

$$\alpha_3 = \rho, \tag{23}$$

$$\alpha_4 = \rho, \tag{24}$$

where $\rho$ is a time-specified gait, which will be specified in Section 4.

**Proposition 3.** When the roll angle and pitch angle of the tilt-rotor are close to zero, the cat-trot gait introduces no singular problem.

**Proof**

Substituting (21) – (24) into the left side of (20) yields

$$4\cos^2(\rho). \tag{25}$$

Since $\rho$ lies in $[-0.65, 0.65]$, (25) is non-zero, meeting Condition (20).

3.4. Third-order PD Controller

Design the third-order PD controllers as

$$\begin{bmatrix}\ddddot{y}_{1d}\\ \ddddot{y}_{2d}\\ \ddddot{y}_{3d}\end{bmatrix} = \begin{bmatrix}\ddddot{y}_{1r}\\ \ddddot{y}_{2r}\\ \ddddot{y}_{3r}\end{bmatrix} + K_{P1}{}^{3\times 3} \cdot \left(\begin{bmatrix}\ddot{y}_{1r}\\ \ddot{y}_{2r}\\ \ddot{y}_{3r}\end{bmatrix}-\begin{bmatrix}\ddot{y}_{1}\\ \ddot{y}_{2}\\ \ddot{y}_{3}\end{bmatrix}\right) + K_{P2}{}^{3\times 3} \cdot \left(\begin{bmatrix}\dot{y}_{1r}\\ \dot{y}_{2r}\\ \dot{y}_{3r}\end{bmatrix}-\begin{bmatrix}\dot{y}_{1}\\ \dot{y}_{2}\\ \dot{y}_{3}\end{bmatrix}\right) + K_{P3}{}^{3\times 3} \cdot \left(\begin{bmatrix}y_{1r}\\ y_{2r}\\ y_{3r}\end{bmatrix}-\begin{bmatrix}y_{1}\\ y_{2}\\ y_{3}\end{bmatrix}\right), \tag{26}$$

$$\ddddot{y}_{4d} = \ddddot{y}_{4r} + K_{PZ_1} \cdot (\ddot{y}_{4r}-\ddot{y}_4) + K_{PZ_2} \cdot (\dot{y}_{4r}-\dot{y}_4) + K_{PZ_3} \cdot (y_{4r}-y_4), \tag{27}$$

where $K_{Pi}$ ($i = 1,2,3$) is the 3-by-3 diagonal control coefficient matrix, $K_{PZ_i}$ ($i = 1,2,3$) is the control coefficient (scalar), $y_j$ ($j = 1,2,3,4$) is the state, $y_{jr}$ ($j = 1,2,3,4$) is the reference.

The control parameters in the above controller are specified as: $K_{P1} = \text{diag}([1,1,1])$, $K_{P2} = \text{diag}([10,10,1])$, $K_{P3} = \text{diag}([50,50,1])$, $K_{PZ_1} = 10$, $K_{PZ_2} = 5$, $K_{PZ_3} = 10$.

As for tracking the rest degrees of freedoms ($X$ and $Y$), firstly design the PD controller

$$\begin{bmatrix}\ddot{X}_d\\ \ddot{Y}_d\end{bmatrix} = \begin{bmatrix}\ddot{X}_r\\ \ddot{Y}_r\end{bmatrix} + \begin{bmatrix}K_{X_1} & 0\\ 0 & K_{Y_1}\end{bmatrix} \cdot \left(\begin{bmatrix}\dot{X}_r\\ \dot{Y}_r\end{bmatrix}-\begin{bmatrix}\dot{X}\\ \dot{Y}\end{bmatrix}\right) + \begin{bmatrix}K_{X_2} & 0\\ 0 & K_{Y_2}\end{bmatrix} \cdot \left(\begin{bmatrix}X_r\\ Y_r\end{bmatrix}-\begin{bmatrix}X\\ Y\end{bmatrix}\right) \tag{28}$$

where $K_{X_1} = K_{Y_1} = K_{X_2} = K_{Y_2} = 1$.

The output (left side) of this PD controller is the desired position, which will be transferred to the desired attitude by modified attitude-position decoupler,

$$\phi_r = \frac{1}{g} \cdot (\ddot{X}_d \cdot s\psi_r - \ddot{Y}_d \cdot c\psi_r) + \frac{F_Y}{mg}, \tag{29}$$

$$\theta_r = \frac{1}{g} \cdot (\ddot{X}_d \cdot c\psi_r + \ddot{Y}_d \cdot s\psi_r) - \frac{F_X}{mg}. \tag{30}$$

where $\psi_r$ is the reference of yaw angle, which is kept as zero in this research.

Notice that the attitude-altitude controller in (26) is designed much faster than the position tracking controller in (28).

## 4. Cat-trot-inspired Gait

This research adopts three different gaits (fixed tilting angles, cat trot with instant switch, and cat trot with continuous switch.

### 4.1. Fixed Tilting Angles

The gait with the fixed tilting angles is defined by

$$\alpha_1 = \alpha_2 = -\rho, \tag{31}$$

$$\alpha_1 = \alpha_2 = \rho. \tag{32}$$

The tilting angles remains constant ($\rho\ remains\ constant$) during the entire flight. In the simulations, we compare the results with different $\rho$ (0.65, 0.65/2, 0, -0.65/2, -0.65).

### 4.2. Cat Trot with Instant Switch

Typical Cat gaits can be categorized as walk gait [30], trot gait [31,32], and gallop gait [31,32], which is determined by the current velocity of a cat.

Parallel to the trot gait of a cat, we set the tilting angle 0.65 radian at most for out tilt-rotor. This gait is specified as

$$\begin{cases} t - n \cdot T \in \left[0, \frac{T}{2}\right] : \rho = 0.65 \\ t - n \cdot T \in \left(\frac{T}{2}, T\right) : \rho = -0.65 \end{cases} \tag{33}$$

where $\rho$ determines the tilting angles in (21) – (24), $T$ is the period of the gait, $t$ represents the current time, $n$ is the floor of $\frac{t}{T}$. Note that the tilting angles change instantly in this gait plan. Thus, we call this cat-trot gait with instant switch.

We compare the result in the simulations with different periods.

### 4.3. Cat Trot with Continuous Switch

In the cat-trot gait with instant switch, the tilting angles changes discretely. This section provides a continuous cat-trot gait,

$$\begin{cases} t - n \cdot T \in \left[0, \frac{T}{2}\right] : \rho = 0.65 - 2 \times 0.65 \cdot \frac{2}{T} \cdot (t - n \cdot T) \\ t - n \cdot T \in \left(\frac{T}{2}, T\right) : \rho = -0.65 + 2 \times 0.65 \cdot \frac{2}{T} \cdot \left(t - n \cdot T - \frac{T}{2}\right) \end{cases} \tag{34}$$

where $\rho$ determines the tilting angles in (21) – (24), $T$ is the period of the gait, $t$ represents the current time, $n$ is the floor of $\frac{t}{T}$.

Also, we compare the result in the simulations with different periods.

## 5. Stability Analysis

In this section, we address the discussion on the stability of our controller. Firstly, we provide the stability proof to our controller. Secondly, we make comments on the potential state errors.

### 5.1. Stability Proof

Noticing that the singular decoupling matrix is avoided (see Proposition 3), the original nonlinear system is stable if the linearized system is proved stable, given that the constraints are not activated; advanced stability analyses are demanded if the constraints are activated [39]. In this research, these constraints are not activated, which simplifies our discussion on the stability.

Proposition 4. (Exponential Stability of Attitude and Altitude) The attitude and altitude are exponentially stable if

$$\begin{bmatrix} K_{P1} & \\ & K_{PZ_1} \end{bmatrix} > 0, \tag{35}$$

$$\begin{bmatrix} K_{P2} & \\ & K_{PZ_2} \end{bmatrix} > 0, \tag{36}$$

$$\begin{bmatrix} K_{P3} & \\ & K_{PZ_3} \end{bmatrix} > 0, \tag{37}$$

$$\begin{bmatrix} K_{P1} & \\ & K_{PZ_1} \end{bmatrix} \cdot \begin{bmatrix} K_{P2} & \\ & K_{PZ_2} \end{bmatrix} - \begin{bmatrix} K_{P3} & \\ & K_{PZ_3} \end{bmatrix} > 0. \tag{38}$$

Proof. Applying Hurwitz Criterion or Routh Criterion to (26) and (27) yields (35) – (38).

Proposition 5. (Exponential Stability of Position $(X,Y)$) The position $(X,Y)$ is exponentially stable if

$$\begin{bmatrix} K_{X_1} & 0 \\ 0 & K_{Y_1} \end{bmatrix} > 0, \tag{39}$$

$$\begin{bmatrix} K_{X_2} & 0 \\ 0 & K_{Y_2} \end{bmatrix} > 0. \tag{40}$$

Proof. Applying Hurwitz Criterion or Routh Criterion to (28) yields (39) – (40).

Thus, all degrees of freedom have been proved exponentially stable.

### 5.2. Remarks to The Stability Proof

In Section 5.1, we proved that the attitude and altitude are exponentially stable, which was verified by simulations in [21], where the attitude and altitude became exponentially stable. However, the reference in [21] were attitude and altitude. The position $(X,Y)$ is not required to be stabilized.

Actually, there can be steady state error in position using this controller.

Remark 6. (Steady State Error in Position $(X,Y)$ for tilt-rotors) There can be steady state error while apply the controllers in (26) – (30). This is because that the modified attitude-position decoupler are deduced by linearization at the state defined in (12) – (14). Consequently, (29) and (30) are precise at the state in (12) – (14) only.

However, maintaining this state produces the acceleration along $X$ or $Y$ directions for a tilt-rotor, which is not generally expected by the reference. As the consequence, the system may be stabilized at a state near the state for linearization (blue point in Figure 3). The unprecise relationship between the position and the attitude can introduce unexpected steady state error.

Remark 7. (Steady State Error in Position $(X,Y)$ for conventional quadrotors) There are no steady state errors for a conventional quadrotor applying PD controllers in a position-tracking problem [40,41]. The equilibrium state for linearization to decouple the conventional quadrotor's attitude and position is also at (12) – (14).

However, a conventional quadrotor produces no lateral accelerations while maintaining this state, which is expected by the reference in general. Stabilizing at the state for linearization gives a precise relationship in attitude and position (red point in Figure 3). Consequently, no steady state error is introduced in the conventional quadrotor.

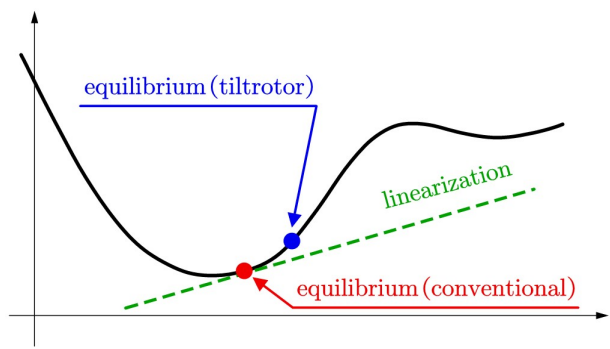

Figure 3. The linearization bias in the tilt-rotor.

6. Trajectory-tracking Experiments

The tilt-rotor is expected to track three types of trajectories in our experiments. They are setpoint, uniform rectilinear motion, and uniform circular motion.

6.1. Setpoint

One may notice that [21] tracks the setpoint using the similar controller. However, the setpoint set in [21] is the attitude-altitude reference (e.g., $(\phi,\theta,\psi,Z) = (0,0,0,0)$). The position $(X,Y)$ is not stabilized; stabilizing at $(\phi,\theta,\psi,Z) = (0,0,0,0)$ produces accelerations along $X$ and $Y$ for the tilt-rotor.

In modified attitude-position decoupler (Proposition 1), we explicit the relationship between the position $(X,Y)$ and attitude, providing the possibility of position-tracking for the tilt-rotor.

The setpoint reference designed in this experiment is position-yaw pair,

$$(X,Y,Z,\psi) = (0,0,0,0) \tag{41}$$

We track this set point adopting the gait with the fixed tilting angles $\rho$ (0.65, 0.65/2, 0, -0.65/2, -0.65) (Section 4.1).

The steady state errors are compared in the results of the simulations with different fixed tilting angles.

We also record the steady state errors for the cases while utilizing the conventional attitude-position decoupler (7), (8) instead of the modified attitude-position decoupler in Proposition 1.

6.2. Uniform Rectilinear Motion

The thorough introductions on the cat gaits can be referred to [30–32], where the typical speed of a walk gait for a cat is $0.54 - 0.74\ m/s$ [30], the typical speeds of a trot gait and gallop gait for a cat are $2.12\ m/s$ and $3.5\ m/s$, respectively [31].

Since we adopted the cat-trot gait in gait plan, we pair it with our reference in (42), whose speed is around $2.12\ m/s$.

$$\begin{cases} X_r = 1.5 \cdot t \\ Y_r = 1.5 \cdot t \end{cases}. \tag{42}$$

The fixed tilting angle gaits ($\rho = 0.65$) in Section 4.1, the cat trot with instant switch with different periods in Section 4.2, the cat trot with continuous switch with different periods in Section 4.3 are adopted in tracking this trajectory (42).

In comparison, we also display the results for the cases where the conventional attitude-position decoupler is adopted.

6.3. Uniform Circular Motion

A circular trajectory (uniform circular motion) is designed as

$$\begin{cases} X_r = 5 \cdot \cos(0.1 \cdot t) \\ Y_r = 5 \cdot \sin(0.1 \cdot t) \end{cases}, \tag{43}$$

$$\begin{cases} \dot{X}_r = -0.1 \times 5 \cdot \sin(0.1 \cdot t) \\ \dot{Y}_r = 0.1 \times 5 \cdot \cos(0.1 \cdot t) \end{cases}. \tag{44}$$

The radius of the trajectory is 5. The period is $20\pi$.

Due to the potential bias discussed in Remark 6, the acceleration of this circular is not feedforwarded while tracking.

The fixed tilting angle gaits ($\rho = 0.65$) in Section 4.1, the cat trot with instant switch with different periods in Section 4.2, the cat trot with continuous switch with different periods in Section 4.3 are adopted in tracking this trajectory (43).

In comparison, we also display the results for the cases where the conventional attitude-position decoupler is adopted.

6.4. Initial Condition

The absolute value of each initial angular velocity of the propellers is 300 rad/s. This angular velocity is not sufficient to compensate the effect of the gravity and will cause unstable in attitude if maintaining this speed.

7. Results

This section displays the results of the tracking experiments with different references.

7.1. Setpoint

The setpoint are tracked by the gaits with the conventional attitude-position decoupler (marked as $B$ in Figure 4) and with the modified attitude-position decoupler (marked as $A$ in Figure 4).

The steady state errors along $X$ and $Y$ ($e_X, e_Y$) are reported in Figure 4.

$A_1(\rho=0.65)$, $A_2\left(\rho=\frac{0.65}{2}\right)$, $A_3(\rho=0)$, $A_4\left(\rho=-\frac{0.65}{2}\right)$, $A_5(\rho=-0.65)$ in Figure 4 represent the tracking results (steady state error) applying the relevant gait equipped with the conventional attitude-position decoupler. Similarly, $B_1(\rho=0.65)$, $B_2\left(\rho=\frac{0.65}{2}\right)$, $B_3(\rho=0)$, $B_4\left(\rho=-\frac{0.65}{2}\right)$, $B_5(\rho=-0.65)$ in Figure 4 represent the tracking results (steady state error) applying the relevant gait equipped with the modified attitude-position decoupler.

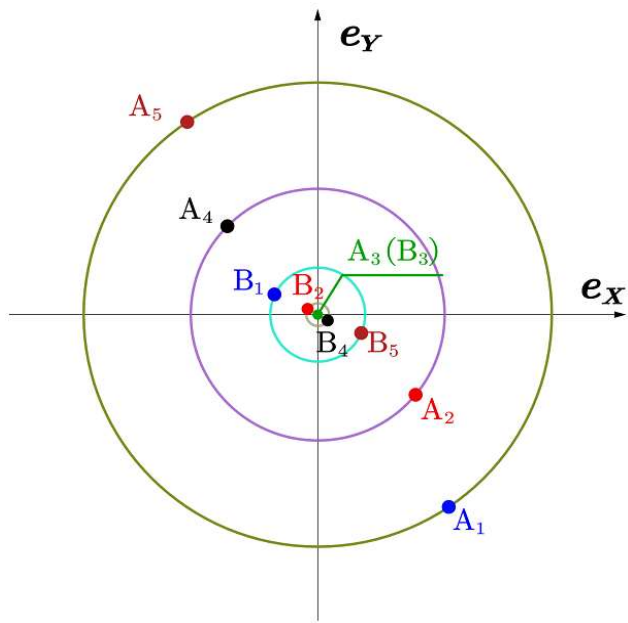

Figure 4. The steady state errors. $A_1(3.35,-3.56)$, $A_2(1.61,-1.64)$, $A_3(0,0)$, $A_4(-1.61,1.64)$, $A_5(-3.35,3.56)$, $B_1(-0.38,0.17)$, $B_2(-0.04,0.02)$, $B_3(0,0)$, $B_4(0.04,-0.02)$, $B_5(0.38,-0.17)$.

Notice that the gait $\rho = 0$ receives zero dynamic state error for both cases with the conventional attitude-position decoupler (marked as $B_3$) and with the modified attitude-position decoupler (marked as $A_3$). This is because that the tilt-rotor degrades to the conventional quadrotor in this gait, which produces no lateral forces (Remark 7). For the rest gaits, the modified attitude-position decoupler notably reduces the steady state error in the same gait.

7.2. Uniform Rectilinear Motion

Figure 5 displays the dynamic state errors for the gait $\rho = 0.65$ with the conventional attitude-position decoupler and with the modified attitude-position decoupler. It can be clearly seen that the steady state errors are reduced after equipping with the modified attitude-position decoupler.

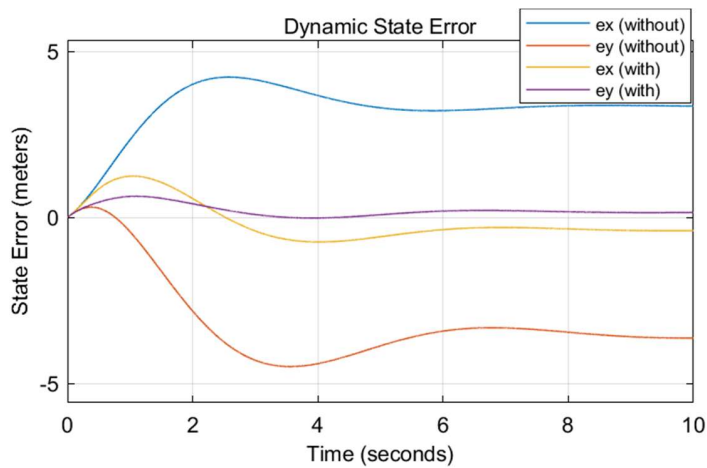

Figure 5. Dynamic state errors ($\rho = 0.65$) in the simulations equipped with the conventional attitude-position decoupler (yellow and purple curves) and with the modified attitude-position decoupler (blue and red curves).

The reduced dynamic state error is also observed in the cat-trot gait ($T = 2s$), applying with the modified attitude-position decoupler, with instant switch (Figure 6) and with continuous switch (Figure 7).

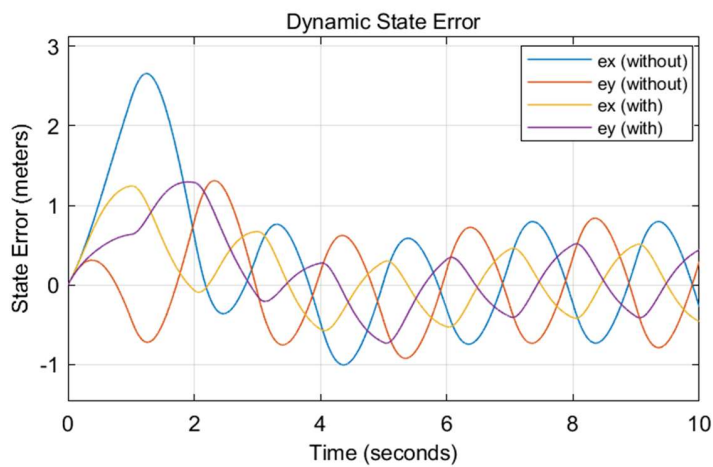

Figure 6. Dynamic state errors in the simulation adopting the cat-trot gait with instant switch($T = 2s$) equipped with the conventional attitude-position decoupler (yellow and purple curves) and with the modified attitude-position decoupler (blue and red curves).

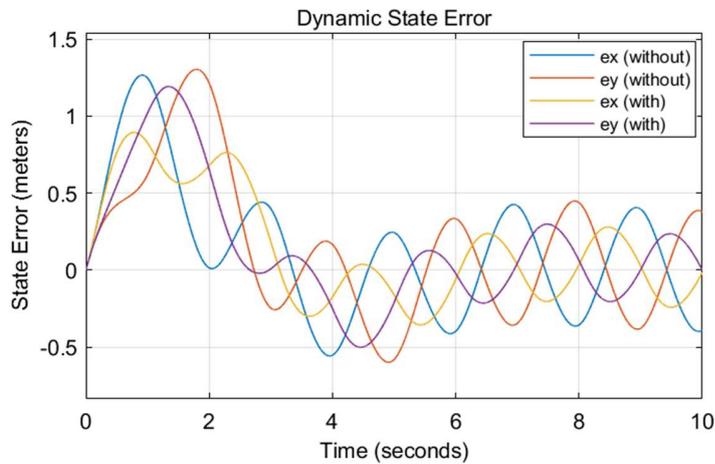

Figure 7. Dynamic state errors in the simulation adopting the cat-trot gait with continuous switch($T = 2s$) equipped with the conventional attitude-position decoupler (yellow and purple curves) and with the modified attitude-position decoupler (blue and red curves).

The relationship between the supremum of the dynamic state error after sufficiently long time and the length of the period of the cat-trot gait with instant switch and cat-trot gait with continuous switch are plotted in Figure 8 and 9, respectively. The red curve represents the result of the simulation equipped with the conventional attitude-position decoupler. While the blue curve represents the result of the simulation equipped with the modified attitude-position decoupler.

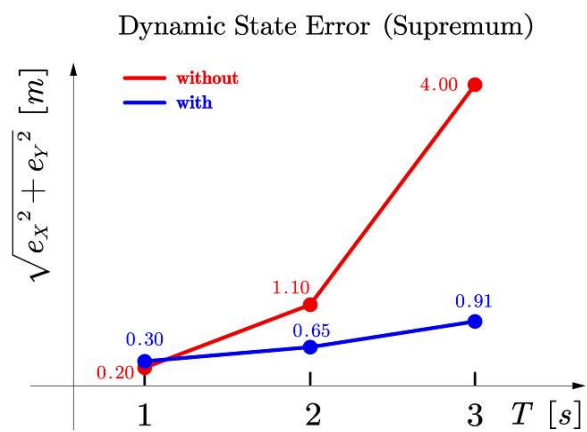

Figure 8. Supremum of the dynamic state error after sufficiently long time in cat-trot gait with instant switch.

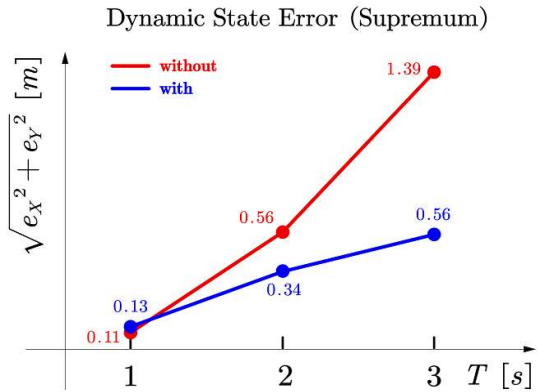

Figure 9. Supremum of the dynamic state error after sufficiently long time in cat-trot gait with continuous switch.

Clearly, a gait with a longer period results in larger dynamic state error. The modified attitude-position decoupler significantly reduces the dynamic state error especially for the gait with long period. While no significant difference in the dynamic state error is reported for the cases with the conventional attitude-position decoupler and with the modified attitude-position decoupler when the period is short.

In addition, the cat-trot gait with continuous switch results in less dynamic state error comparing with the cat-trot gait with instant switch, given the identical period and decoupler.

7.3. Uniform Circular Motion

Figure 10 displays the dynamic state errors for the fixed-tilting-angle gait $\rho = 0.65$ with the conventional attitude-position decoupler and with the modified attitude-position decoupler in tracking the uniform circular reference. Clearly, the modified attitude-position decoupler receives less dynamic state error comparing with the conventional attitude-position decoupler.

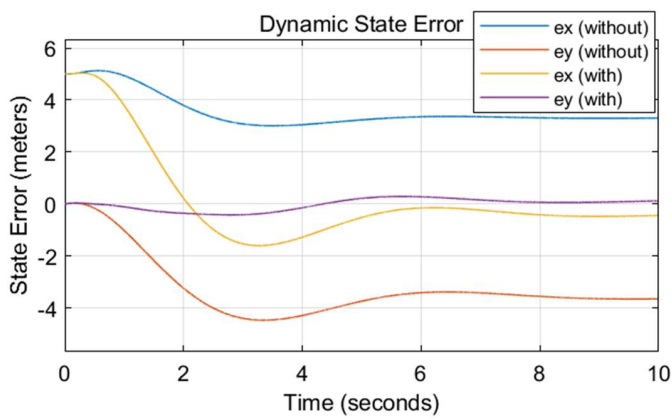

Figure 10. Dynamic state errors ($\rho = 0.65$) in the simulations equipped with the conventional attitude-position decoupler (yellow and purple curves) and with the modified attitude-position decoupler (blue and red curves).

Likewise, the reduced dynamic state error is also observed in the cat-trot gait ($T = 2s$), applying with the modified attitude-position decoupler, with instant switch (Figure 11) and with continuous switch (Figure 12).

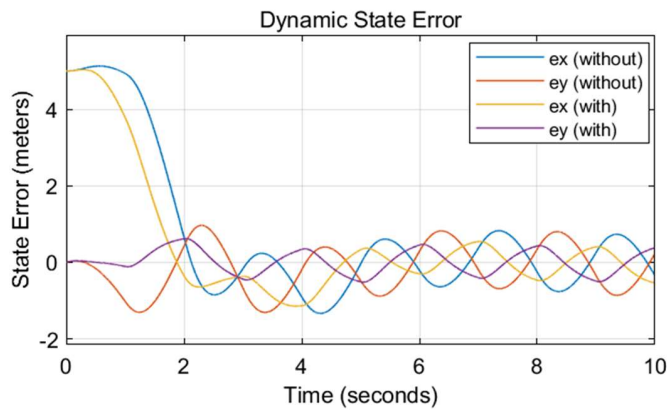

Figure 11. Dynamic state errors in the simulation adopting the cat-trot gait with instant switch($T = 2s$) equipped with the conventional attitude-position decoupler (yellow and purple curves) and with the modified attitude-position decoupler (blue and red curves).

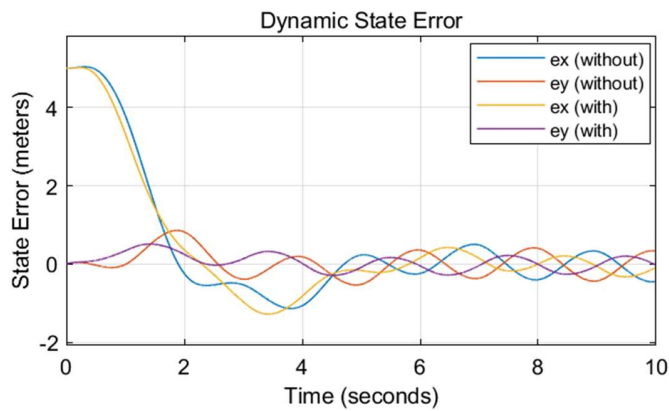

Figure 12. Dynamic state errors in the simulation adopting the cat-trot gait with continuous switch($T = 2s$) equipped with the conventional attitude-position decoupler (yellow and purple curves) and with the modified attitude-position decoupler (blue and red curves).

The relationship between the supremum of the dynamic state error after sufficiently long time and the length of the period of the cat-trot gait with instant switch and cat-trot gait with continuous switch are plotted in Figure 13 and 14, respectively. The red curve represents the result of the simulation equipped with the conventional attitude-position decoupler. While the blue curve represents the result of the simulation equipped with the modified attitude-position decoupler.

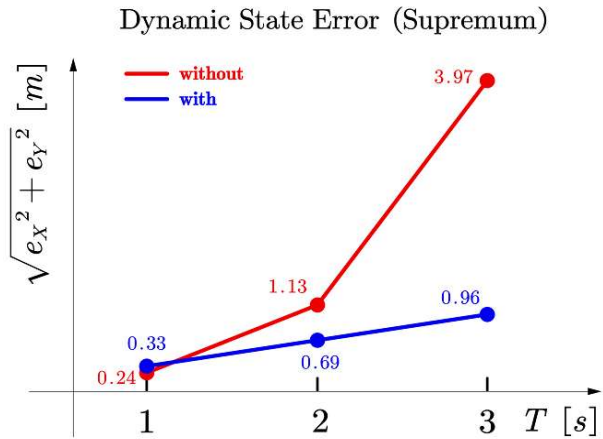

Figure 13. Supremum of the dynamic state error after sufficiently long time in cat-trot gait with instant switch.

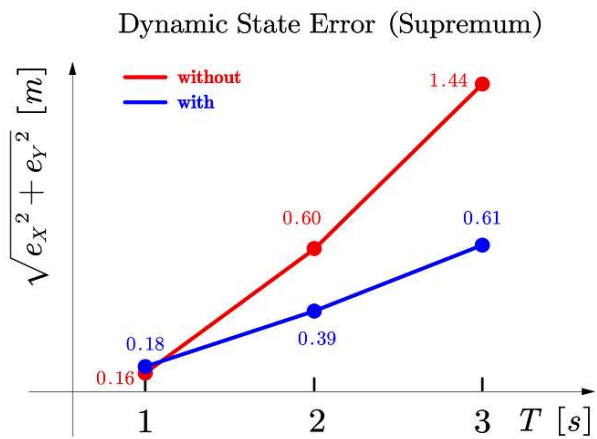

Figure 14. Supremum of the dynamic state error after sufficiently long time in cat-trot gait with continuous switch.

Clearly, a gait with a longer period results in larger dynamic state error. The modified attitude-position decoupler significantly reduces the dynamic state error especially for the gait with long period. While no significant difference in the dynamic state error is reported for the cases with the conventional attitude-position decoupler and with the modified attitude-position decoupler when the period is short.

In addition, the cat-trot gait with continuous switch results in less dynamic state error comparing with the cat-trot gait with instant switch, given the identical period and decoupler.

8. Conclusions and Discussions

The decoupling matrix is invertible in feedback linearization for a tilt-rotor, adopting cat-trot gait. All the references considered in this research (setpoint, uniform rectilinear motion, uniform circular motion) are successfully tracked with acceptable steady state error by cat-trot gait.

The relationship between position and attitude of the tilt-rotor is elucidated. The modified attitude-position decoupler is invented for position-tracking problem for a tilt-rotor. This significantly reduces the dynamic state error or steady state error comparing with the conventional attitude-position decoupler.

The length of the period of the cat-trot gait highly influences the dynamic state error in tracking a uniform rectilinear reference and a uniform circular reference. Specifically, a gait with the shorter the period tends to receive less the supremum of the dynamic state error after sufficiently long time. On the other hand, the modified attitude-position decoupler is particularly powerful in reducing the dynamic state error for a gait with long period.

In general, the continuous cat-trot gait receives the less supremum of the dynamic state error after sufficiently long time than the discrete cat-trot gait, given the identical period and decoupler.

There are several questions remaining to be explored.

Firstly, Equation (17) takes the tilting angles constant; they are assumed to be constant during the entire flight. While the switching process in the discrete gait or the entire period in the continuous gait violates this condition. Addressing the discussions on the underlying robustness is beyond the scope of this research, which can be a further step.

Secondly, though the steady state error is reported reduced after applying the modified attitude-position decoupler, further reduction of the dynamic state error may be possible by not ignoring the high-order infinitesimal terms while deducing the modified attitude-position decoupler.

Also, analysis on other different cat gaits (e.g., gallop trot and walk trat) can be another further step before adopting for the gait plan for the tilt-rotor.